%% file: main.tex
\documentclass[runningheads]{llncs}

\input{mymacros}
\usepackage[T1]{fontenc}
\usepackage{microtype}

\usepackage{graphicx}
\usepackage{tabularx}
\usepackage{float}
\usepackage[misc]{ifsym}
\usepackage{fontawesome5}
\usepackage{orcidlink}
\begin{document}
\title{Anatomy-Structured Hierarchical MIL for Weakly-Supervised Thoracic Disease Detection in Chest X-rays}
\titlerunning{ASH-MIL}

\author{Jeongin Kim\inst{1}\orcidlink{0009-0007-0535-5547} \and
Sohyun Ahn\inst{2}\orcidlink{0000-0002-0116-3325} \and
Seo Young Kang\inst{3}\orcidlink{0000-0003-2431-3397} \and
Jaeyi Sung\inst{1}\orcidlink{0009-0007-4587-3146} \and \\
Soomin Kim\inst{1}\orcidlink{0009-0005-8983-6203} \and
Sungho Cho\inst{4}\orcidlink{0009-0002-0246-2957} \and
Rena Lee\inst{4}\orcidlink{0009-0006-2992-6802} \and 
Kwanchang Kim\inst{5}\textsuperscript{(\Letter)}\orcidlink{0000-0001-8297-5415} \and 
Junhyug Noh\inst{1}\textsuperscript{(\Letter)}\orcidlink{0000-0003-1239-8178}
}

\index{Kim, Jeongin}
\index{Ahn, Sohyun}
\index{Kang, Seo Young}
\index{Sung, Jaeyi}
\index{Kim, Soomin}
\index{Cho, Sungho}
\index{Lee, Rena}
\index{Kim, Kwanchang}
\index{Noh, Junhyug}

\authorrunning{J. Kim et al.}
\institute{
Division of Artificial Intelligence \& Software, Ewha Womans University \and
Ewha Medical Artificial Intelligence Research Institute, Ewha Womans University \and
Department of Nuclear Medicine, Ewha Womans University \and
REMEDI Inc. R\&D Center \and
Ewha Womans University Seoul Hospital \\
\email{mdkkchang@ewha.ac.kr}, \quad
\email{junhyug@ewha.ac.kr}
}

\maketitle
\begin{abstract}
Weakly-supervised thoracic disease detection in chest X-rays (CXR) is challenging due to subtle appearances and complex anatomical overlap, motivating anatomy-aware modeling for improved localization. However, prior anatomy-aware methods typically rely on coarse region proxies or static spatial priors, which may restrict dynamic instance discovery and limit precise localization of small abnormalities. We propose Anatomy-Structured Hierarchical Multiple Instance Learning (ASH-MIL), a framework that introduces parallel anatomy-structured observation branches (\emph{cardiac}, \emph{pulmonary}, and \emph{agnostic}) combined with hierarchical MIL aggregation. Anatomical priors are injected as soft spatial biases into decoder cross-attention, enabling anatomically grounded evidence maps without disease bounding-box supervision.
Instance localization is derived directly from MIL-weighted cross-attention maps without bounding box supervision. Experiments on CXR8 and cross-domain MIMIC-CXR held-out sets demonstrate consistent improvements over prior weakly-supervised and anatomy-aware approaches, particularly under stricter localization criteria.
Our code is available at \href{https://github.com/jn-kim/ash-mil.git}{\faGithub\ \texttt{jn-kim/ash-mil}}.
\keywords{Weakly-Supervised Object Detection, Multiple Instance Learning, Anatomical Priors, Chest X-ray}
\end{abstract}

\section{Introduction}
\label{sec:intro}

Chest radiography is the most frequently performed diagnostic imaging exam worldwide, accounting for a large fraction of clinical imaging volume~\cite{ccalli2021deep}. Despite its ubiquity, accurate interpretation remains challenging: radiologist error rates are estimated at 3--5\% in routine practice and can be substantially higher in abnormal-only settings~\cite{gefter2023commonly}. A major contributor is perceptual misses, which are particularly common for subtle findings in radiographic blind spots~\cite{gefter2023commonly,brady2017error,klein2019systematic}. Improving lesion localization may therefore help reduce missed findings and enhance diagnostic reliability.

However, obtaining instance-level bounding boxes in chest X-rays (CXR) is labor-intensive, motivating weakly-supervised object detection (WSOD) using only image-level labels. Many CXR WSOD methods rely on class activation maps (CAM) or global attention to highlight discriminative regions~\cite{chexnet,wsrpn,vmamba,nih_cxr8,zhou2016learning,gradcam}. While effective for post-hoc visualization, these unconstrained global search mechanisms often yield anatomically implausible evidence and cannot explicitly restrict the model’s focus to clinically relevant organs, complicating integration into clinical workflows.

In contrast, radiologists follow an anatomically structured search strategy, scanning predefined landmarks and evaluating region-specific patterns before forming a diagnosis~\cite{klein2019systematic,kundel1972visual,van2017visual,kelly2016development}. Existing anatomy-aware approaches incorporate anatomy via static coordinate priors that ignore patient-specific morphology~\cite{thorax-priornet} or via auxiliary feature-level guidance from proxy tasks~\cite{agxnet}, but they do not restructure the underlying instance discovery process.

We propose \textbf{Anatomy-Structured Hierarchical Multiple Instance Learning (ASH-MIL)}, which explicitly organizes weakly-supervised thoracic disease detection under anatomical priors. ASH-MIL introduces three parallel anatomy-structured observation branches (\emph{cardiac}, \emph{pulmonary}, \emph{agnostic}) that act as complementary anatomical observers. Each branch predicts the full label space while being softly guided by patient-specific organ priors injected as additive cross-attention biases in a query-based decoder. To learn from image-level supervision, we formulate detection as hierarchical multiple instance learning: we first aggregate query-level hypotheses within each branch, then fuse branch-level evidence with class-dependent attention. Instance localization is derived directly from MIL-weighted cross-attention maps, avoiding bounding box supervision.

Our contributions are:
\begin{itemize}
\item We introduce an anatomy-structured WSOD framework built on a CXR-pretrained RAD-DINO backbone~\cite{rad_dino}, injecting organ-level priors as soft attention biases into learnable object queries.
\item We propose a hierarchical MIL scheme for query- and branch-level evidence aggregation, enabling anatomically grounded evidence attribution and localization from cross-attention under image-level labels.
\item We conduct comprehensive experiments and ablations on CXR8~\cite{nih_cxr8} and MIMIC-CXR~\cite{mimic-cxr-heldout}, demonstrating consistent gains, particularly for small and subtle abnormalities.
\end{itemize}

\section{Related Work}

\noindent \textbf{Weakly-supervised CXR detection.}
Most WSOD methods for chest X-rays rely on CAM-style localization under image-level labels~\cite{chexnet,nih_cxr8}. Proposal- or pooling-based variants such as WSRPN~\cite{wsrpn} improve localization quality, but they remain largely anatomy-agnostic and do not structure instance discovery by anatomical context.

\noindent \textbf{Anatomy-aware CXR interpretation.}
Anatomical cues have been incorporated via region-level supervision or attention transfer, \eg ADPD~\cite{adpd} and AGXNet~\cite{agxnet}, or via coordinate priors such as ThoraX-PriorNet~\cite{thorax-priornet}. These approaches typically rely on coarse/static priors that can be sensitive to patient-specific morphology. In contrast, ASH-MIL injects organ-based priors as soft cross-attention biases in specialized branches and combines them via hierarchical MIL, enabling flexible anatomy-conditioned reasoning.

\section{Method}
\label{sec:method}

The proposed ASH-MIL framework addresses weakly-supervised thoracic disease detection under anatomical priors by introducing three parallel anatomy-structured observation branches -- \emph{cardiac}, \emph{pulmonary}, and \emph{agnostic}. Each branch predicts
the full label space, while anatomical specialization emerges via soft spatial priors injected into the
transformer decoder. To learn from image-level supervision only, we formulate detection as a hierarchical MIL
problem: (i) query-level aggregation (QLA) within each branch and (ii) branch-level aggregation (BLA) across branches.

\subsection{Problem Setting}
\label{subsec:problem}

We formulate the task as WSOD, where multiple disease instances may exist for each class within a single image.
Given an input CXR image $I$, the model is trained using only image-level labels $\mathbf{y} \in \{0,1\}^C$, indicating the presence or absence of each of the $C$ disease categories.
The objective is to predict class-wise confidence scores and derive corresponding bounding boxes $b = [x, y, w, h]$ for multiple disease instances without instance-level annotations.

\begin{figure*}[t!]
\centering
\includegraphics[width=\textwidth]{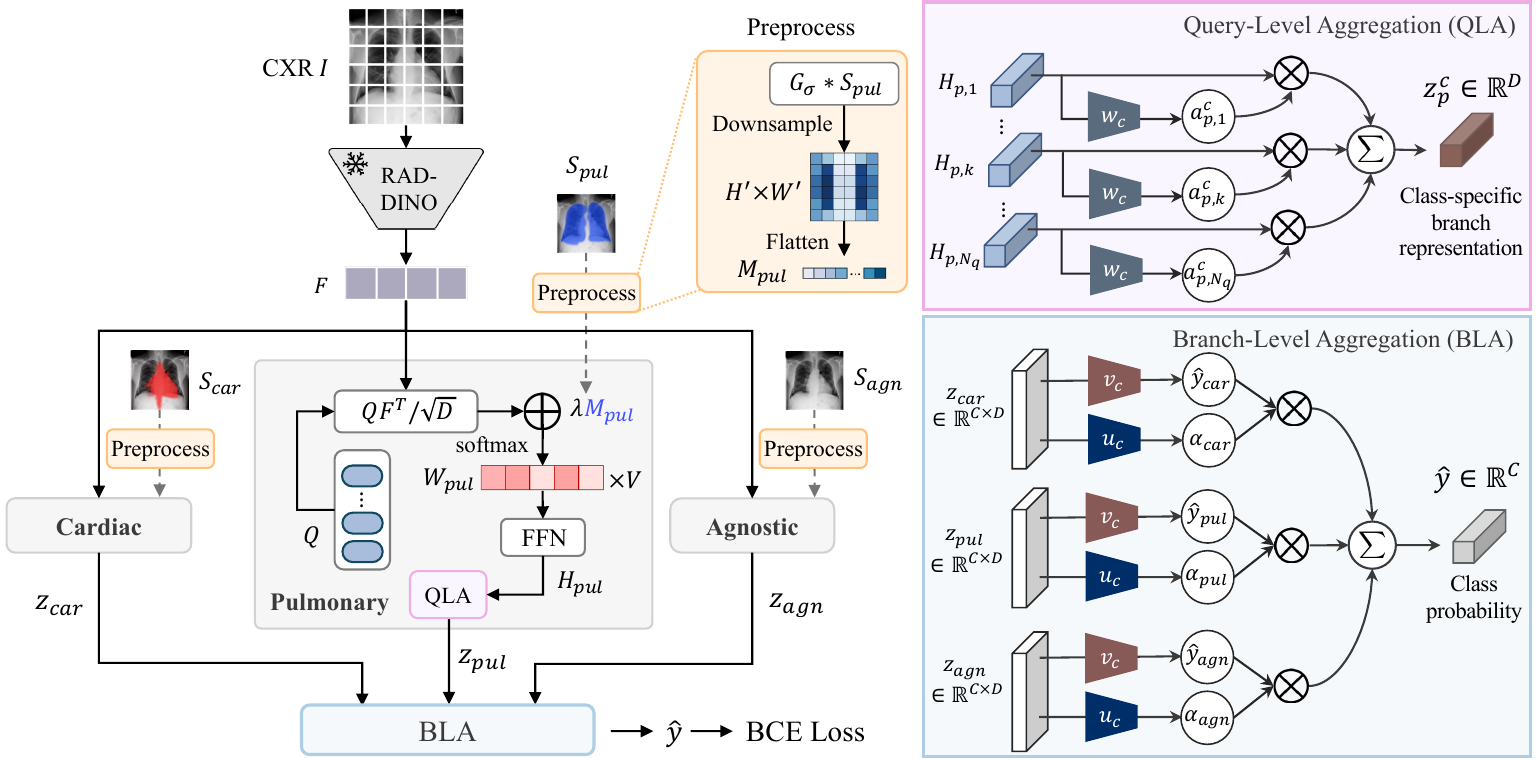}
\caption{
\textbf{Overall architecture of ASH-MIL.}
A shared RAD-DINO backbone extracts patch tokens that are processed by three parallel anatomy-structured decoder branches: cardiac (\textit{car}), pulmonary (\textit{pul}), and agnostic (\textit{agn}). For brevity, only the pulmonary branch is illustrated in detail. The right panel presents the hierarchical MIL framework that aggregates query-level and branch-level evidence to produce the final class-wise predictions.
}
\label{fig:method_overview}
\end{figure*}

\subsection{Model}
\label{subsec:arch}

Fig.~\ref{fig:method_overview} provides an overview of ASH-MIL. We design three parallel observation branches as complementary anatomical observers: \emph{cardiac}, \emph{pulmonary}, and \emph{agnostic}. We leverage RAD-DINO~\cite{rad_dino}, a self-supervised ViT~\cite{vit} pre-trained on chest X-rays via DINOv2~\cite{dinov2}, as our shared backbone $\mathcal{E}$. RAD-DINO captures dense, high-fidelity radiographic representations that preserve fine-grained visual details often underrepresented in textual supervision from radiology reports. This yields a robust, domain-specific foundation for identifying subtle thoracic abnormalities. Given an input image $I$, we extract spatially contextualized patch tokens $\boldsymbol{F} = \mathcal{E}(I) \in \mathbb{R}^{T \times D}$, where $T$ and $D$ denote the number of spatial tokens and the embedding dimension, respectively.

\smallskip
\noindent\textbf{Anatomical priors.}
We construct branch-specific spatial priors $M_p \in \mathbb{R}^{1\times T}$ using CXAS~\cite{cxas}, an off-the-shelf anatomy segmentation model that provides fine-grained pixel-level masks for $157$ chest anatomical structures.
Here, $p \in \mathcal{P}=\{\mathit{car},\,\mathit{pul},\,\mathit{agn}\}$ denote the cardiac, pulmonary, and agnostic branches, respectively. We define clinically relevant subsets $\mathcal{A}_p$ by selecting a subset of CXAS structures for each branch, with $\mathcal{A}_{\mathit{car}}=\{\text{heart, cardiomediastinum, aorta, etc.}\}$ and $\mathcal{A}_{\mathit{pul}}=\{\text{lung}\}$, while $\mathcal{A}_{\mathit{agn}}=\emptyset$.

For each branch $p$, we first aggregate anatomical masks via pixel-wise union, $S_p = \bigcup_{n\in\mathcal{A}_p} s_n$. To align the anatomical prior with the spatial token grid of the backbone, we transform $S_p$ via Gaussian smoothing, bilinear downsampling, and flattening to obtain $M_p$:
\begin{equation}
M_p = \mathrm{Flatten}(\mathrm{Bilinear}(G_\sigma \ast S_p)) \in \mathbb{R}^{1 \times T}
\end{equation}
where $T = H'W'$ denotes the total number of spatial tokens.

\smallskip
\noindent\textbf{Anatomy-biased decoder branch.}
Following the query-based decoding paradigm of DETR~\cite{detr}, we adopt a decoder-only transformer architecture. In DETR, a transformer encoder is used to globally contextualize CNN features. Here, our RAD-DINO ViT backbone yields globally contextualized tokens via stacked self-attention layers, allowing us to omit the additional encoder and directly use $\boldsymbol{F}$ in the decoder branches. For each branch $p$, we employ an independent decoder with $N_q$ learnable queries $Q_p \in \mathbb{R}^{N_q\times D}$. The anatomical prior $M_p$ is incorporated as an additive bias to the cross-attention logits, scaled by $\lambda$:
\begin{equation}
W_p = \mathrm{softmax}_t \left( \frac{Q_p \boldsymbol{F}^\top}{\sqrt{D}} + \lambda M_p \right) \in \mathbb{R}^{N_q\times T},
\end{equation}
where $M_p$ is broadcast across queries and $t$ denotes the spatial patch token index ($t \in \{1,\dots,T\}$). The cross-attention output $\tilde{H}_p = W_p V$, where $V$ denotes the linearly projected backbone tokens, is then processed through a standard transformer decoder block to obtain the refined query embeddings. 
Each block follows the conventional structure with residual connections and LayerNorm after both the attention and feed-forward layers. 
The refined query representations are denoted as
\begin{equation}
H_p = \mathrm{FFN}(\tilde{H}_p)
= [H_{p,1}, \dots, H_{p,N_q}]^\top \in \mathbb{R}^{N_q\times D},
\end{equation}
where $H_{p,k} \in \mathbb{R}^{D}$ serves as the $k$-th instance for the subsequent hierarchical MIL pooling.

\subsection{Training}
\label{subsec:mil}

\noindent\textbf{Hierarchical aggregation.}
Given the branch-specific query embeddings $H_p = [H_{p,1}, \dots, H_{p,N_q}]^\top$, 
we perform query-level aggregation (QLA) within each branch, followed by branch-level aggregation (BLA) across branches (Fig.~\ref{fig:method_overview}, right).

For branch $p$ and class $c$, a class-specific representation is computed as 
$z_p^c = \sum_{k=1}^{N_q} a_{p,k}^c H_{p,k} \in \mathbb{R}^D$, 
where the attention weights 
$a_{p,k}^c = \mathrm{softmax}_k(w_c^\top H_{p,k})$ 
measure the relevance of each query embedding to a learnable vector $w_c \in \mathbb{R}^D$. 
The resulting branch-specific prediction is 
$\hat{y}_p^c = \sigma(v_c^\top z_p^c)$, 
with $v_c \in \mathbb{R}^D$ denoting the class classifier. 

We then aggregate branch-specific predictions via BLA by computing class-dependent branch attention
$\alpha_p^c=\mathrm{softmax}_p(u_c^\top z_p^c)$ and obtain the final prediction as
\begin{equation}
\hat{y}^c = \sum_{p \in \mathcal{P}} \alpha_p^c \hat{y}_p^c,
\end{equation}
where $\mathcal{P}=\{\mathit{car},\mathit{pul},\mathit{agn}\}$ denotes the set of branches and
$u_c \in \mathbb{R}^D$ is a learnable branch-scoring vector.

\smallskip
\noindent\textbf{Training objective.}
We optimize the multi-label binary cross-entropy loss. 
Because instance-level annotations are unavailable, we dispense with the bipartite Hungarian matching standard in DETR~\cite{detr}. 
Instead, our hierarchical MIL formulation provides supervision by dynamically routing gradients to informative queries and anatomical branches directly from the image-level objective.

\subsection{Inference}
\label{subsec:inference}

At inference time, the final image-level prediction for class $c$ is
$\hat{y}^c = \sum_{p \in \mathcal{P}} \alpha_p^c \hat{y}_p^c$,
where $\alpha_p^c$ and $\hat{y}_p^c$ denote the branch-level attention weights and branch-specific predictions, respectively.
We derive bounding boxes from the decoder cross-attention and hierarchical MIL weights. For class $c$, we aggregate token-level evidence as
\begin{equation}
E^c(t)
= \sum_{p \in \mathcal{P}}
\alpha_p^c
\sum_{k=1}^{N_q} a_{p,k}^c [W_p]_{k,t}.
\end{equation}
The resulting token map is reshaped and upsampled to the image resolution. 
High-evidence regions are identified by selecting the minimal pixel set whose cumulative mass exceeds $\tau=0.8$, and these regions are grouped into spatially connected components to extract tight bounding boxes. 
Each box is assigned a confidence score $s^c = \hat{y}^c \cdot \left( \sum_{u \in R} E^c(u) / \sum_{u} E^c(u) \right)$, which reflects both the image-level prediction and the evidence mass within the region $R$.

\section{Experiments}
\label{sec:exp}

\subsection{Experimental Setup}
\label{subsec:setup}

\noindent\textbf{Datasets.}
We primarily train the models on NIH ChestX-ray8 (CXR8)~\cite{nih_cxr8} ($107{,}010$ images, $8$ disease classes),
with patient-level $80\%/10\%/10\%$ train/val/test splits.
All weakly-supervised methods are trained using only image-level labels.
For localization on CXR8, we use the expert-annotated subset of $880$ images with $984$ boxes, split into
$432/448$ images for validation/test.
We further evaluate cross-domain generalization on the MIMIC-CXR held-out set~\cite{mimic-cxr-heldout,mimic-cxr-jpg}
($354$ images, $458$ boxes: $234$ pneumonia, $224$ pneumothorax).\footnote{\url{https://github.com/leotam/MIMIC-CXR-annotations}}
All datasets are publicly available and de-identified, and we used them in accordance with their respective licenses and data use agreements; no additional IRB approval or patient consent was required for this study.

\smallskip
\noindent\textbf{Baselines.}
We compare against anatomy-aware methods (ADPD~\cite{adpd}, ThoraX-PriorNet~\cite{thorax-priornet}, AGXNet~\cite{agxnet})
and anatomy-agnostic weakly-supervised baselines (WSRPN~\cite{wsrpn}, CheXNet~\cite{chexnet}, Grad-CAM~\cite{gradcam},
CXR~\cite{nih_cxr8}).
ADPD and AGXNet rely on additional MIMIC-derived annotations (Chest ImaGenome~\cite{chest-imagenome} and
RadGraph~\cite{radgraph}) unavailable in CXR8; thus, we evaluate ADPD on MIMIC without finetuning, and pretrain AGXNet on
MIMIC before finetuning on CXR8.
For CheXNet and CAM-based methods, boxes are obtained by thresholding activation maps and extracting connected components.
We also include a fully supervised DETR~\cite{detr} with an ImageNet-pretrained ResNet-101 backbone, trained on
the annotated CXR8 validation images and evaluated on the test images.

\smallskip
\noindent\textbf{Implementation details.}
Images are resized to $512\times512$. We use a CXR-pretrained RAD-DINO backbone%
\footnote{\url{https://huggingface.co/microsoft/rad-dino}} with a DETR-style decoder and learnable object queries.
Training uses a single RTX A6000 GPU (peak memory $\sim$18\,GB), batch size $32$, and random horizontal flip ($p=0.5$).
We optimize with AdamW (lr $2\times10^{-4}$, weight decay $10^{-4}$), apply gradient clipping ($0.1$), and use StepLR for
$50$ epochs with early stopping (patience $=5$).

\smallskip
\noindent\textbf{Evaluation metrics.}
We report AP (\%) and CorLoc~\cite{corloc} at IoU thresholds $\{0.1,0.3,0.5\}$, and their means over IoU $0.1$--$0.5$ (step $0.1$).

\begin{table}[t!]
\caption{Detection performance on CXR8 and MIMIC-CXR datasets.}
\label{tab:tab1}
\centering
\fontsize{8}{9}\selectfont
\setlength{\tabcolsep}{3.0pt}
\begin{tabular}{l c c c c c c c c}
\toprule
& \multicolumn{2}{c}{IoU@0.1} & \multicolumn{2}{c}{IoU@0.3} & \multicolumn{2}{c}{IoU@0.5} & \multicolumn{2}{c}{IoU@0.1--0.5} \\
\cmidrule(lr){2-3} \cmidrule(lr){4-5} \cmidrule(lr){6-7} \cmidrule(lr){8-9}
Method & AP & CorLoc & AP & CorLoc & AP & CorLoc & mAP & CorLoc \\
\midrule

\multicolumn{9}{c}{\textbf{CXR8}} \\
\midrule
\textbf{ASH-MIL (Ours)} & \textbf{28.82} & \textbf{0.81} & \textbf{14.23} & \textbf{0.59} & \textbf{6.03} & \textbf{0.26} & \textbf{15.77} & \textbf{0.55} \\
\addlinespace[0.5pt]
\multicolumn{9}{l}{\emph{Ablations}}\\
\quad 3 branch (w/o prior) & 25.02 & 0.68 & 11.48 & 0.44 & 2.05 & 0.18 & 13.07 & 0.43 \\
\quad 1 branch (w/ prior)  & 22.55 & 0.60 &  8.86 & 0.32 & 3.31 & 0.13 & 11.35 & 0.34 \\
\quad 1 branch (w/o prior) & 24.66 & 0.68 & 10.81 & 0.40 & 2.23 & 0.14 & 12.12 & 0.40 \\
\quad Mean pooling         &  3.89 & 0.22 &  1.70 & 0.12 & 0.76 & 0.05 &  2.00 & 0.13 \\
\addlinespace[0.5pt]
\multicolumn{9}{l}{\emph{Anatomy-aware baselines}}\\
\quad ADPD~\cite{adpd} &  6.58 & 0.38 & 2.00 & 0.16 & 0.56 & 0.05 &  3.05 & 0.19 \\
\quad ThoraX-PriorNet~\cite{thorax-priornet} & 10.15 & 0.65 & 4.92 & 0.34 & 0.03 & 0.09 & 5.11 & 0.35 \\
\quad AGXNet~\cite{agxnet} &  5.44 & 0.43 & 1.28 & 0.15 & 0.48 & 0.05 &  2.29 & 0.21 \\
\addlinespace[0.5pt]
\multicolumn{9}{l}{\emph{WSOD / CAM baselines}}\\
\quad WSRPN~\cite{wsrpn} & 10.34 & 0.21 & 6.44 & 0.12 & 3.21 & 0.06 &  6.67 & 0.16 \\
\quad CheXNet~\cite{chexnet} & 24.28 & 0.56 & 12.25 & 0.32 & 3.89 & 0.11 &  12.97 & 0.33 \\
\quad Grad-CAM~\cite{gradcam} & 10.15 & 0.37 & 2.13 & 0.14 & 0.20 & 0.03 &  4.16 & 0.18 \\
\quad ChestX-ray8~\cite{nih_cxr8} & 10.16 & 0.58 & 8.51 & 0.27 & 3.50 & 0.19 & 4.72 & 0.21 \\
\addlinespace[0.5pt]
\multicolumn{9}{l}{\emph{Fully supervised}}\\
\quad DETR~\cite{detr} & 12.49 & 0.49 & 10.67 & 0.32 & 5.30 & 0.25 &  8.81 & 0.33 \\
\midrule

\multicolumn{9}{c}{\textbf{MIMIC-CXR}} \\
\midrule
\textbf{ASH-MIL (Ours)} & \textbf{36.12} & \textbf{0.66} & \textbf{24.53} & \textbf{0.50} & \textbf{7.28} & \textbf{0.25} & \textbf{22.72} & \textbf{0.48} \\
\addlinespace[0.5pt]
\multicolumn{9}{l}{\emph{Ablations}}\\
\quad 3 branch (w/o prior) & 33.76 & 0.56 & 22.01 & 0.47 & 7.12 & 0.22 & 22.26 & 0.43 \\
\quad 1 branch (w/ prior)  & 27.52 & 0.54 & 21.44 & 0.45 & 6.37 & 0.23 & 18.89 & 0.42 \\
\quad 1 branch (w/o prior) & 29.24 & 0.59 & 17.35 & 0.44 & 4.29 & 0.18 & 16.66 & 0.41 \\
\quad Mean pooling         &  8.10 & 0.22 &  4.55 & 0.14 & 1.96 & 0.05 &  4.62 & 0.13 \\
\addlinespace[0.5pt]
\multicolumn{9}{l}{\emph{Anatomy-aware baselines}}\\
\quad ADPD~\cite{adpd} & 11.89 & 0.36 & 0.81 & 0.06 & 0.03 & 0.01 &  4.24 & 0.14 \\
\quad ThoraX-PriorNet~\cite{thorax-priornet} & 26.23 & 0.63 & 15.69 & 0.40 & 5.52 & 0.19 & 15.47 & 0.41 \\
\quad AGXNet~\cite{agxnet} & 19.36 & 0.66 &  9.58 & 0.44 & 1.93 & 0.17 & 10.70 & 0.45 \\
\addlinespace[0.5pt]
\multicolumn{9}{l}{\emph{WSOD / CAM baselines}}\\
\quad WSRPN~\cite{wsrpn} & 10.23 & 0.35 & 5.36 & 0.24 & 2.14 & 0.13 & 11.44 & 0.28 \\
\quad CheXNet~\cite{chexnet} & 22.90 & 0.45 & 12.59 & 0.29 & 3.88 & 0.14 & 12.92 & 0.29 \\
\quad Grad-CAM~\cite{gradcam} &  8.86 & 0.20 &  2.09 & 0.08 & 0.14 & 0.01 &  3.70 & 0.09 \\
\quad ChestX-ray8~\cite{nih_cxr8} & 4.17 & 0.11 & 2.06 & 0.07 & 1.58 & 0.01 & 3.00 & 0.05 \\
\addlinespace[0.5pt]
\multicolumn{9}{l}{\emph{Fully supervised}}\\
\quad DETR~\cite{detr} &  9.52 & 0.57 &  5.43 & 0.36 & 3.80 & 0.22 &  6.00 & 0.36 \\
\bottomrule
\end{tabular}
\end{table}

\subsection{Experimental Results}
\label{subsec:results}

\noindent\textbf{Main results on CXR8 and cross-domain MIMIC-CXR.}
Table~\ref{tab:tab1} reports localization performance on CXR8 and MIMIC-CXR held-out set.
ASH-MIL consistently outperforms prior weakly-supervised and anatomy-aware baselines across all IoU thresholds on both benchmarks.
The gains are more pronounced under stricter criteria (IoU@0.3 and IoU@0.5), indicating more precise instance localization beyond coarse CAM/attention-based cues.
On CXR8, ASH-MIL achieves \textbf{15.77} mAP (IoU@0.1--0.5) and improves AP/CorLoc at IoU@0.5 to \textbf{6.03}/\textbf{0.26}.
On MIMIC-CXR, it attains \textbf{22.72} mAP and maintains strong localization at higher IoU thresholds, demonstrating robust cross-domain generalization.

\smallskip
\noindent\textbf{Component analysis.}
We analyze four ablations in Table~\ref{tab:tab1} to quantify the effects of anatomical priors, the multi-branch design, and hierarchical MIL.
Removing the prior from the 3-branch model (3 branch, w/o prior) reduces performance, particularly under strict criteria
(\eg CXR8 AP/CorLoc@0.5: 6.03/0.26 $\rightarrow$ 2.05/0.18), confirming that soft spatial biasing improves precise localization.
Collapsing the architecture to a single branch (1 branch, w/ prior) underperforms the corresponding prior-free variant (1 branch, w/o prior),
suggesting that merging heterogeneous anatomical contexts into a single observer can dilute context-specific cues and hinder attention focusing.
In contrast, the full 3-branch design with priors yields the best performance
(CXR8 mAP: 15.77 vs.\ 12.12 for 1 branch, w/o prior; MIMIC mAP: 22.72 vs.\ 16.66).
Finally, replacing hierarchical MIL with mean pooling causes a drastic drop, demonstrating that attention-based query/branch aggregation is critical for routing
gradients to informative hypotheses under image-level supervision.

\begin{figure}[t!]
\centering
\includegraphics[width=0.8\textwidth]{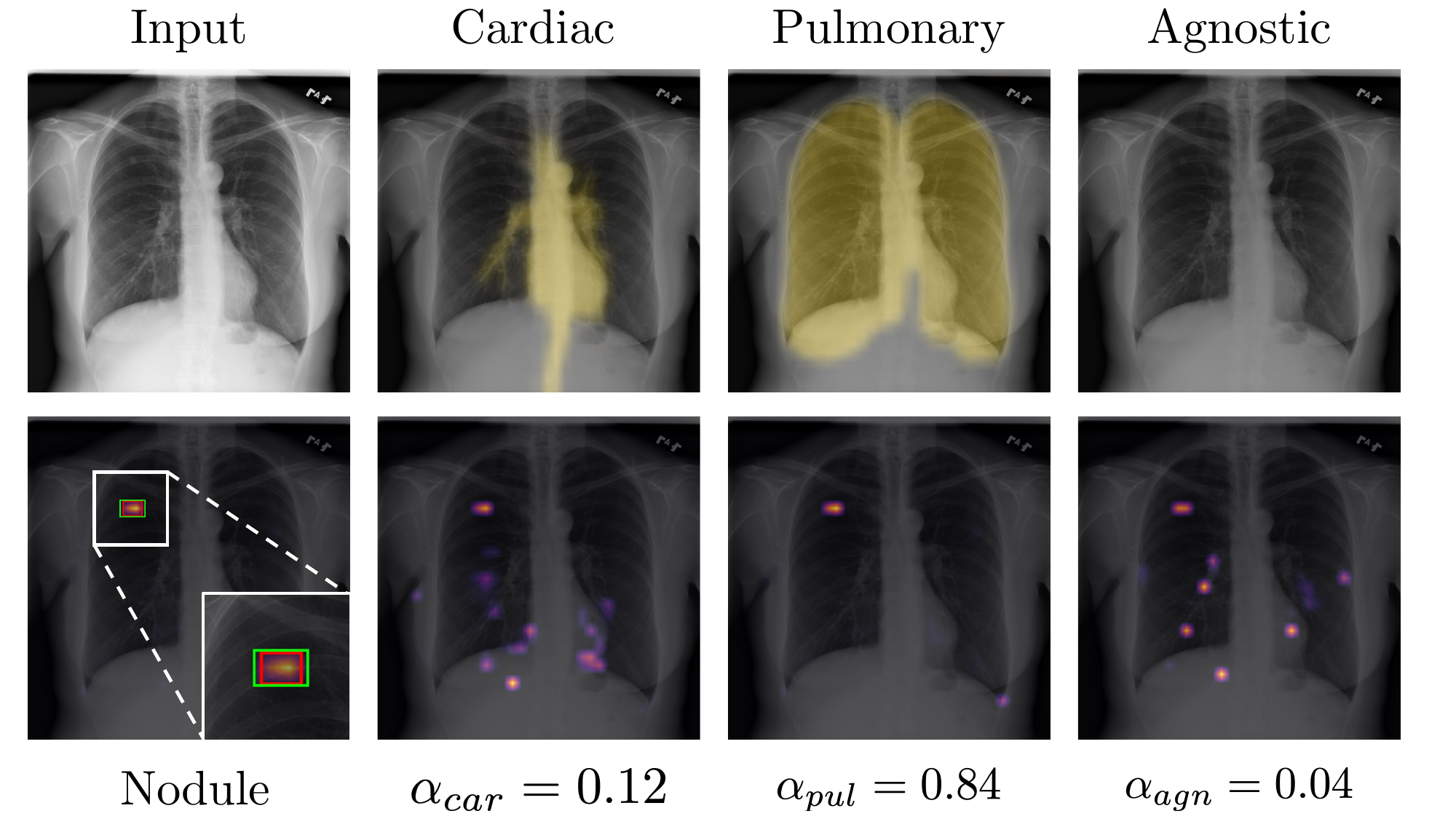}
\caption{Qualitative results on CXR8. \textbf{Top:} the input image and branch-specific anatomical priors (cardiac, pulmonary, agnostic). \textbf{Bottom:} the final Nodule localization (GT: red; Pred: green) and the corresponding branch evidence maps. The learned branch attention weights $\alpha_p$ quantify each branch's contribution to the Nodule prediction.}
\label{fig:qual}
\end{figure}

\smallskip
\noindent\textbf{Qualitative evidence routing.}
Fig.~\ref{fig:qual} illustrates how ASH-MIL selects anatomically relevant observers.
For a small nodule within the lung parenchyma, the pulmonary branch receives the highest attention weight
($\alpha_{\textit{pul}}=0.84$), and its evidence map is spatially aligned with the lesion.
The fused evidence is therefore dominated by the pulmonary observer, producing a compact localization that closely matches the ground truth.

\section{Conclusion}
\label{sec:conc}

We present \textbf{ASH-MIL}, an anatomy-structured weakly supervised framework for thoracic disease detection in chest X-rays. ASH-MIL introduces parallel anatomy-conditioned observers (cardiac, pulmonary, agnostic) by injecting organ priors as soft cross-attention biases in a query-based decoder, and learns instance discovery from image-level labels via hierarchical MIL that aggregates evidence across queries and branches. Experiments on CXR8 and cross-domain MIMIC-CXR show consistent localization gains, especially for small and subtle abnormalities, while providing interpretable branch-level evidence routing; limitations include dependence on automated anatomy segmentation quality and a fixed coarse branch design for boundary-spanning findings.

\begin{credits}
\subsubsection{\ackname}
This work was supported by the National Research Foundation of Korea (NRF) grants funded by the Korean government (MSIT) (RS-2026-25499022 and RS-2025-25394457), the G-LAMP Program of the NRF funded by the Ministry of Education (RS-2025-25442252), the Basic Science Research Program through the NRF funded by the Ministry of Education (RS-2026-25472409), and the Korea Institute for Advancement of Technology (KIAT) grant funded by the Korean government (Ministry of Trade, Industry and Energy) (P0030108, Industrial Technology Innovation Project -- International Joint Technology Development Project -- Strategic Technology Type -- Global Demand-Linked Type).

\subsubsection{\discintname} The authors have no competing interests to declare that are relevant to the content of this article.
\end{credits}
%
%
%
%
\bibliographystyle{splncs04}
\bibliography{references}

\setcounter{table}{0}

\end{document}

%% file: mymacros.tex
\usepackage{color}
\usepackage{algorithm}
\usepackage{algpseudocode}
\usepackage{amsmath,amssymb,amsfonts}
\usepackage{booktabs}       
\usepackage{multirow, varwidth}
\usepackage{mathtools}
\usepackage{epsfig}
\usepackage{xr}
\usepackage{flushend}
\usepackage{url}
\usepackage{enumitem}
\setlist{nosep} 
\usepackage{xcolor}         
\usepackage{kotex}
\usepackage{colortbl}
\usepackage{hyperref}
\usepackage{comment}

\DeclareRobustCommand\onedot{\futurelet\@let@token\@onedot}
\def\onedot{.} 
\def\eg{\emph{e.g}\onedot, }

\newcommand{\mytilde}{\raise.17ex\hbox{$\scriptstyle\mathtt{\sim}$}}

\definecolor{dark2green}{rgb}{0.1, 0.65, 0.3}

\newif\ifcolormode
\colormodetrue

\newcommand{\setcolormodeon}{\colormodetrue}

\setcolormodeon 

%% file: main.bbl
\begin{thebibliography}{10}
\providecommand{\url}[1]{\texttt{#1}}
\providecommand{\urlprefix}{URL }
\providecommand{\doi}[1]{https://doi.org/#1}

\bibitem{brady2017error}
Brady, A.P.: Error and discrepancy in radiology: inevitable or avoidable? Insights into imaging  \textbf{8}(1),  171--182 (2017)

\bibitem{ccalli2021deep}
{\c{C}}all{\i}, E., Sogancioglu, E., Van~Ginneken, B., Van~Leeuwen, K.G., Murphy, K.: Deep learning for chest x-ray analysis: A survey. Medical image analysis  \textbf{72},  102125 (2021)

\bibitem{detr}
Carion, N., Massa, F., Synnaeve, G., Usunier, N., Kirillov, A., Zagoruyko, S.: End-to-end object detection with transformers. In: Proceedings of the European Conference on Computer Vision. pp. 213--229 (2020)

\bibitem{corloc}
Deselaers, T., Alexe, B., Ferrari, V.: Weakly supervised localization and learning with generic knowledge. International Journal of Computer Vision  \textbf{100}(3),  275--293 (2012)

\bibitem{vit}
Dosovitskiy, A., Beyer, L., Kolesnikov, A., Weissenborn, D., Zhai, X., Unterthiner, T., Dehghani, M., Minderer, M., Heigold, G., Gelly, S., et~al.: An image is worth 16x16 words: Transformers for image recognition at scale. In: International Conference on Learning Representations (2021)

\bibitem{gefter2023commonly}
Gefter, W.B., Post, B.A., Hatabu, H.: Commonly missed findings on chest radiographs: causes and consequences. Chest  \textbf{163}(3),  650--661 (2023)

\bibitem{van2017visual}
Van~der Gijp, A., Ravesloot, C., Jarodzka, H., Van~der Schaaf, M., Van~der Schaaf, I., van Schaik, J.P., Ten~Cate, T.J.: How visual search relates to visual diagnostic performance: a narrative systematic review of eye-tracking research in radiology. Advances in Health Sciences Education  \textbf{22}(3),  765--787 (2017)

\bibitem{thorax-priornet}
Hossain, M.I., Zunaed, M., Ahmed, M.K., Hossain, S.J., Hasan, A., Hasan, T.: Thorax-priornet: A novel attention-based architecture using anatomical prior probability maps for thoracic disease classification. IEEE Access  \textbf{12},  3256--3273 (2023)

\bibitem{radgraph}
Jain, S., Agrawal, A., Saporta, A., Truong, S.Q., Duong, D.N., Bui, T., Chambon, P., Zhang, Y., Lungren, M.P., Ng, A.Y., et~al.: Radgraph: Extracting clinical entities and relations from radiology reports. arXiv preprint arXiv:2106.14463  (2021)

\bibitem{mimic-cxr-jpg}
Johnson, A., Pollard, T., Greenbaum, N., Lungren, M., Deng, C., Peng, Y., Lu, Z., Mark, R., Berkowitz, S., Horng, S.: Mimic-cxr-jpg -- chest radiographs with structured labels (version 2.0.0). PhysioNet (2019)

\bibitem{kelly2016development}
Kelly, B.S., Rainford, L.A., Darcy, S.P., Kavanagh, E.C., Toomey, R.J.: The development of expertise in radiology: in chest radiograph interpretation,“expert” search pattern may predate “expert” levels of diagnostic accuracy for pneumothorax identification. Radiology  \textbf{280}(1),  252--260 (2016)

\bibitem{klein2019systematic}
Klein, J.S., Rosado-de Christenson, M.L.: A systematic approach to chest radiographic analysis. Diseases of the Chest, Breast, Heart and Vessels 2019-2022: Diagnostic and Interventional Imaging pp. 1--16 (2019)

\bibitem{kundel1972visual}
Kundel, H.L., La~Follette~Jr, P.S.: Visual search patterns and experience with radiological images. Radiology  \textbf{103}(3),  523--528 (1972)

\bibitem{adpd}
M{\"u}ller, P., Meissen, F., Brandt, J., Kaissis, G., Rueckert, D.: Anatomy-driven pathology detection on chest x-rays. In: Medical Image Computing and Computer-Assisted Intervention. pp. 57--66 (2023)

\bibitem{wsrpn}
M{\"u}ller, P., Meissen, F., Kaissis, G., Rueckert, D.: Weakly supervised object detection in chest x-rays with differentiable roi proposal networks and soft roi pooling. IEEE Transactions on Medical Imaging  (2024)

\bibitem{dinov2}
Oquab, M., Darcet, T., Moutakanni, T., Vo, H., Szafraniec, M., Khalidov, V., Fernandez, P., Haziza, D., Massa, F., El-Nouby, A., et~al.: Dinov2: Learning robust visual features without supervision. arXiv preprint arXiv:2304.07193  (2023)

\bibitem{rad_dino}
P{\'e}rez-Garc{\'\i}a, F., Sharma, H., Bond-Taylor, S., Bouzid, K., et~al.: Exploring scalable medical image encoders beyond text supervision. Nature Machine Intelligence  \textbf{7}(1),  119--130 (2025)

\bibitem{chexnet}
Rajpurkar, P., Irvin, J., Zhu, K., Yang, B., Mehta, H., Duan, T., Ding, D., Bagul, A., Langlotz, C., Shpanskaya, K., et~al.: Chexnet: Radiologist-level pneumonia detection on chest x-rays with deep learning. arXiv preprint arXiv:1711.05225  (2017)

\bibitem{cxas}
Seibold, C., Jaus, A., Fink, M., Kim, M., Rei{\ss}, S., Herrmann, K., Kleesiek, J., Stiefelhagen, R.: Accurate fine-grained segmentation of human anatomy in radiographs via volumetric pseudo-labeling. arXiv preprint arXiv:2306.03934  (2023)

\bibitem{gradcam}
Selvaraju, R., Cogswell, M., Das, A., Vedantam, R., Parikh, D., Batra, D.: Grad-cam: Visual explanations from deep networks via gradient-based localization. In: Proceedings of the IEEE International Conference on Computer Vision. pp. 618--626 (2017)

\bibitem{mimic-cxr-heldout}
Tam, L.K., Wang, X., Turkbey, E., Lu, K., Wen, Y., Xu, D.: Weakly supervised one-stage vision and language disease detection using large scale pneumonia and pneumothorax studies. In: International Conference on Medical Image Computing and Computer-Assisted Intervention. pp. 45--55. Springer (2020)

\bibitem{vmamba}
Wang, T., Huang, K., Xu, M., Huang, J.: Weakly supervised chest x-ray abnormality localization with non-linear modulation and foreground control. Scientific Reports  \textbf{14}(1),  29181 (2024)

\bibitem{nih_cxr8}
Wang, X., Peng, Y., Lu, L., Lu, Z., Bagheri, M., Summers, R.: Chestx-ray8: Hospital-scale chest x-ray database and benchmarks on weakly supervised classification and localization of common thorax diseases. In: Proceedings of the IEEE Conference on Computer Vision and Pattern Recognition. pp. 2097--2106 (2017)

\bibitem{chest-imagenome}
Wu, J.T., Agu, N.N., Lourentzou, I., Sharma, A., Paguio, J.A., Yao, J.S., Dee, E.C., Mitchell, W., Kashyap, S., Giovannini, A., et~al.: Chest imagenome dataset for clinical reasoning. arXiv preprint arXiv:2108.00316  (2021)

\bibitem{agxnet}
Yu, K., Ghosh, S., Liu, Z., Deible, C., Batmanghelich, K.: Anatomy-guided weakly supervised abnormality localization in chest x-rays. In: Medical Image Computing and Computer-Assisted Intervention. pp. 658--668 (2022)

\bibitem{zhou2016learning}
Zhou, B., Khosla, A., Lapedriza, A., Oliva, A., Torralba, A.: Learning deep features for discriminative localization. In: Proceedings of the IEEE conference on computer vision and pattern recognition. pp. 2921--2929 (2016)

\end{thebibliography}
